\documentclass[10pt,twocolumn,letterpaper]{article}

\usepackage{iccv}
\usepackage{times}
\usepackage{epsfig}
\usepackage{graphicx}
\usepackage{amsmath}
\usepackage{amssymb}

\usepackage{graphicx}
\usepackage{amsmath}
\usepackage{amssymb}
\usepackage{booktabs}
\usepackage{multirow}
\usepackage{times}
\usepackage{epsfig}
\usepackage{graphicx}
\usepackage{amsmath}            
\usepackage{amssymb}
\usepackage[table,xcdraw]{xcolor}
\usepackage{mathtools}
\usepackage{adjustbox}
\usepackage{gensymb}
\usepackage{indentfirst}
\usepackage{multirow}
\usepackage[T1]{fontenc}
\usepackage{mathrsfs}
\usepackage{booktabs}
\usepackage{array}
\usepackage{colortbl}
\usepackage{bm}
\usepackage{lipsum}
\usepackage{color}
\usepackage[accsupp]{axessibility}

\usepackage[pagebackref=true,breaklinks=true,letterpaper=true,colorlinks,bookmarks=false]{hyperref}
\usepackage[capitalize]{cleveref}
\iccvfinalcopy 

\def\iccvPaperID{7464} 

\ificcvfinal\fi

\begin{document}

\title{MEFLUT: Unsupervised 1D Lookup Tables for Multi-exposure Image Fusion}

\author{Ting Jiang\,$^{1}$  \quad
Chuan Wang\,$^1$  \quad
Xinpeng Li\,$^1$   \quad
Ru Li\,$^1$  \quad
Haoqiang Fan\,$^1$ \quad
Shuaicheng Liu\,$^{2,1}$\thanks{Corresponding author}  \\
$^1$\,Megvii Technology\\
$^2$\, University of Electronic Science and Technology of China\\
}

\maketitle
\ificcvfinal\thispagestyle{empty}\fi

\begin{abstract}
   In this paper, we introduce a new approach for high-quality multi-exposure image fusion (MEF). We show that the fusion weights of an exposure can be encoded into a 1D lookup table (LUT), which takes pixel intensity value as input and produces fusion weight as output. We learn one 1D LUT for each exposure, then all the pixels from different exposures can query 1D LUT of that exposure independently for high-quality and efficient fusion. Specifically, to learn these 1D LUTs, we involve attention mechanism in various dimensions including frame, channel and spatial ones into the MEF task so as to bring us significant quality improvement over the state-of-the-art (SOTA). In addition, we collect a new MEF dataset consisting of 960 samples, 155 of which are manually tuned by professionals as ground-truth for evaluation. Our network is trained by this dataset in an unsupervised manner. Extensive experiments are conducted to demonstrate the effectiveness of all the newly proposed components, and results show that our approach outperforms the SOTA in our and another representative dataset SICE, both qualitatively and quantitatively. Moreover, our 1D LUT approach takes less than 4ms to run a 4K image on a PC GPU. Given its high quality, efficiency and robustness, our method has been shipped into millions of Android mobiles across multiple brands world-wide. Code is available at: \href{https://github.com/Hedlen/MEFLUT}{https://github.com/Hedlen/MEFLUT}.
\end{abstract}


\vspace{-5mm}
\section{Introduction}\label{sec:intro}
Dynamic ranges of natural scenes are much wider than those captured by commercial imaging products, causing they are hard to be captured by most digital photography sensors. As a result, high dynamic range (HDR)~\cite{bogoni2000extending,yan2019attention} imaging techniques have attracted considerable interest due to their capability to overcome such limitations. 
Among HDR solutions, multi-exposure image fusion (MEF)~\cite{mertens2007exposure, li2017detail, ram2017deepfuse, li2018multi, ma2019deep, zhang2020ifcnn, xu2020u2fusion, xu2020mef, zhang2020rethinking, jung2020unsupervised, xu2020fusiondn, qu2022transmef, zhang2023iid} provides a cost-effective one, with which plausible images with vivid detail can be generated. MEF has attracted wide attention and there have been many methods~\cite{debevec2008recovering, mertens2009exposure, granados2010optimal, li2012detail, li2013image, ma2017robust, kou2017multi, ma2017multi, li2020fast} available to fuse images with faithful detail and color reproduction. However, these MEF methods use hand-crafted features or transformations so that they usually suffer from robustness if being applied to modified conditions.

\begin{figure}[t]
    \centering
    \includegraphics[width=0.98\linewidth]{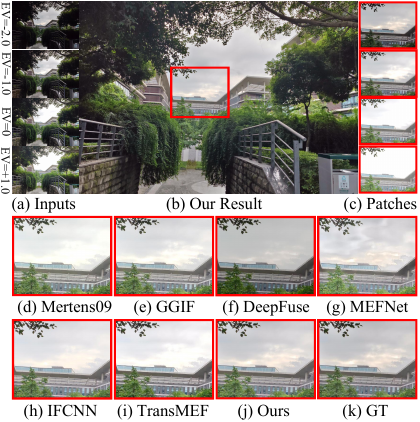}\vspace{-1mm}
    \caption{(a) Inputs. (b) Our result. (c) Input patches. We show the results by various methods in patches. Our result hallucinates more accurate details over the saturated areas.}\vspace{-4mm}
   \label{fig:teaser}
\end{figure}
Inspired by the successes of the deep neural networks (DNNs) in many computer vision areas~\cite{ren2015faster,redmon2018yolov3,szegedy2015going,he2016deep}, recently some deep learning-based approaches ~\cite{ram2017deepfuse,ma2019deep} have been proposed to improve MEF. DeepFuse~\cite{ram2017deepfuse} uses DNNs for the first time to directly regress the Y channel of an YUV image as the target. As the core weight maps' learning is ignored, its quality is limited despite of its acceptable efficiency due to the limited model size. To improve the quality, MEFNet~\cite{ma2019deep} alternatively learns the weight maps for blending the input sequence. However, its speed is downgraded as the network becomes complex, hence it just works on low resolution input as a workaround to maintain the efficiency. 
Unfortunately, these methods do not take real-world deployment into consideration, their speed and quality cannot be well balanced, causing the difficulty of their wide application such as into mobile platform. To tackle these issues, we propose a method named MEFLUT, which aims to achieve higher quality and efficiency simultaneously by taking advantages of deep learning techniques. Our method has no strict requirements on the running platform, it can run in PC and mobile CPU and GPU. As shown in Figs.~\ref{fig:teaser} and~\ref{fig:time}, our method outperforms all the other methods in the image detail preservation and running speed.

MEFLUT consists of two parts. Firstly, we design a network based on multi-dimensional attention mechanism, which is trained via unsupervised methods. The attention mechanism works in frame, channel and spatial dimension separately to fuse inter-frame and intra-frame features, which brings us quality gain in detail preservation. After the network converges, we simplify the model to multiple 1D LUTs, that encodes the fusion weight for a given input pixel value from a specific exposure image. In the test phase, the fusion weights corresponding to different exposures are directly queried from the LUTs. In order to further accelerate our method, the input is downsampled to obtain the fusion masks, which are then upsampled to the original resolution for the fusion, by guided filtering for upsampling (GFU)~\cite{ma2019deep} to avoid boundary stratification. We verify that with or without GFU, our method always runs faster than our competitors, revealing the key effect of LUTs learned.

Besides, considering none of the existing MEF datasets is completely collected through mobile devices, we therefore build a high-quality multi-exposure image sequence dataset. Specifically, we spent over one month to capture and filter out 960 multi-exposure images covering a diversity of scenes by different brands of mobile phones. Among them, 155 samples are produced with ground-truth (GT) for the purpose of quantitative evaluation, which are produced by first running image predictions by 14 algorithms, voted by 20 volunteers and then fine-tuned by an Image Quality (IQ) expert. Producing these GT samples totally cost at least 40 man-hours without counting the organization effort.


To sum up, our main contributions include:
\vspace{-0.2cm}
\begin{itemize}
\setlength\itemsep{-1.5mm}
  
  
  \item We propose MEFLUT that learns 1D LUTs for the task of MEF. We show that the fusion weights can be encoded into the LUTs successfully. Once learned, MEFLUT can be easily deployed with high efficiency, so that a 4K image runs in less than 4ms on a PC GPU. To the best of our knowledge, this is the first time to demonstrate the benefits of LUTs for MEF. 
  
  \item We propose a new network structure with two attention modules in all dimensions that outperforms the state-of-the-art in quality especially detail preservation. 
  
  \item  We also release a new dataset of 960 multi-exposure image sequences collected by mobile phones in various brands and from diverse scenes. 155 samples of them are produced detailed image as ground-truth target by professionals manually.
\end{itemize}

\section{Related Works}\label{sec:related}\vspace{-1mm}



\subsection{Existing MEF Algorithms}
\vspace{-0.5mm}
MEF tasks typically perform as a weighted summation of multiple frames with different exposures.
Therefore, the focus of MEF is often to find an appropriate method to get the weights of different exposures. Mertens~\emph{et al.}~\cite{mertens2009exposure} uses contrast, saturation and well-exposeness of each exposure to get the fused weights. 
Compared with these traditional MEF methods~\cite{goshtasby2005fusion, mertens2007exposure,mertens2009exposure, zhang2011gradient, li2013image, li2017pixel, kou2017multi}, which focus on obtaining the weight in advance, some others~\cite{ram2017deepfuse,endo2017deep, li2020fast, zhang2020ifcnn, xu2020u2fusion, xu2020fusiondn, qu2022transmef, zhang2023iid} prefer to transform the MEF task into an optimization problem.
Ma~\emph{et al.}~\cite{ma2017multi} proposed a gradient based method to minimize MEF-SSIM for search better fusion results in image space. However, this method requires searching in each fusion causing that it is so time-consuming. In recent years, some deep learning methods also try to optimize the model through MEF-SSIM. For example, DeepFuse~\cite{ram2017deepfuse} accomplishes the MEF task via a neural network, achieving faster computation than the traditional method while keeping the fusion quality. Recently, Zhang~\emph{et al.}~\cite{zhang2020ifcnn} proposed a general image fusion framework IFCNN, which is based on DNNs and direct reconstructs the fusion results through the network.  Qu~\emph{et al.}~\cite{qu2022transmef} proposed the TransMEF, which uses a Transformer to further improve the quality of MEF. However, these methods do not consider speed and are not designed for mobile devices.

\begin{figure*}[htbp]
 \vspace{-6mm}
 \centering
  \begin{minipage}[t]{0.95\linewidth}
   \centering
   \includegraphics[width=0.93\linewidth]{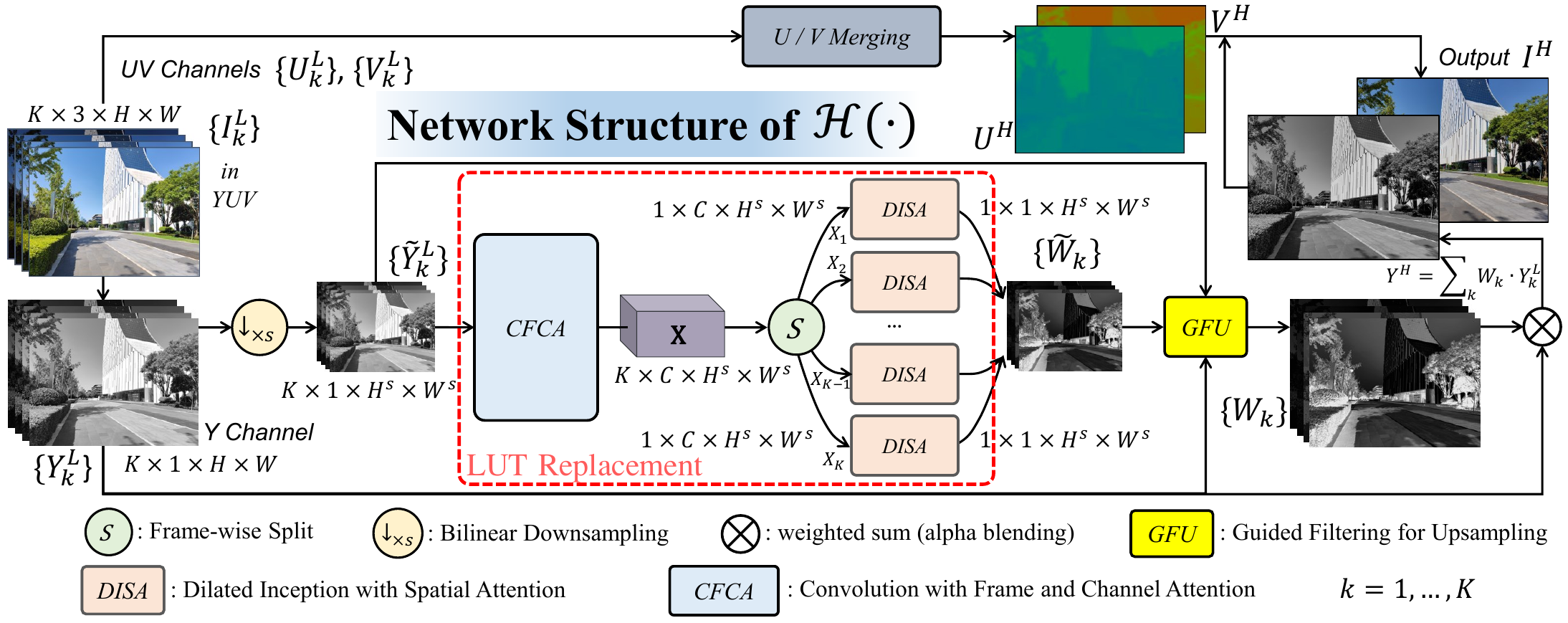}
  \end{minipage}
    \begin{minipage}[t]{0.95\linewidth}
     \centering
     \includegraphics[width=0.93\linewidth]{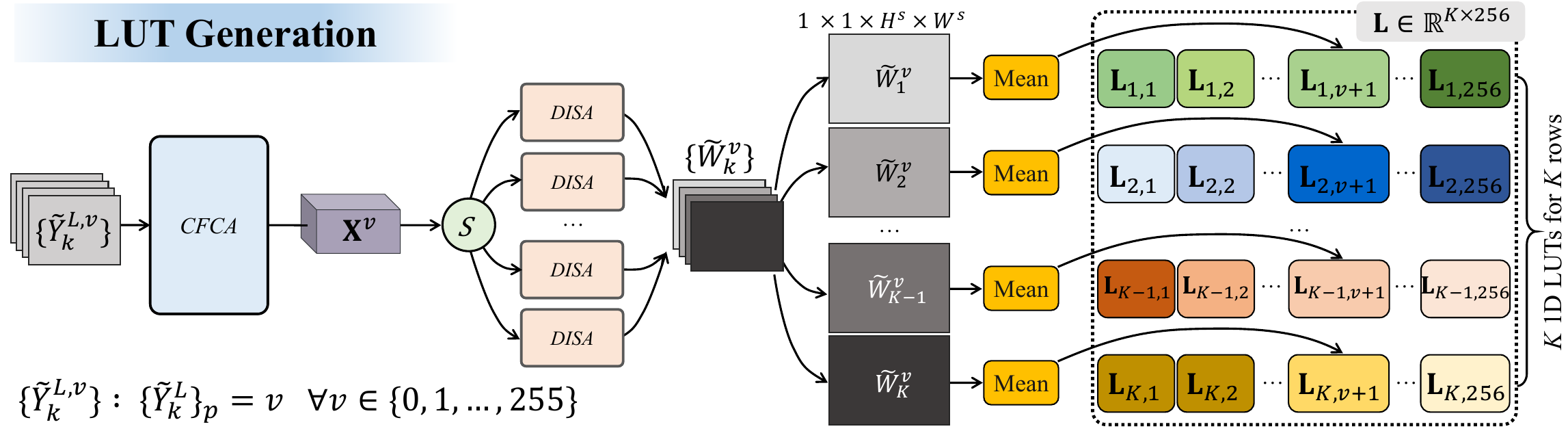}
    \end{minipage}
\caption{\textbf{Top}: Our network structure $\mathcal{H}(\cdotp)$ for unsupervised training. \textbf{Bottom}: Generation of $K$ 1D LUTs. For each grayscale $v$, we feed the network with a group of YUV images with constant color $v$ and set the LUT matrix by the predicted results.} 
\label{fig:pipeline}
\vspace{-3mm}
\end{figure*}
\subsection{Acceleration of MEF Algorithms}
\vspace{-0.5mm}
Most of the aforementioned methods are time-consuming considering potential deployment in mobile platform, as mobile devices are relatively of limited computing power. 
One solution is to cloud-based solution while for high resolution images, image transmission is also time-consuming. Another solution is to conduct the computation in down-sampled version of images like~\cite{burt1987laplacian, burt1993enhanced,hasinoff2016burst,gharbi2017deep,chen2017fast, ma2019deep}. 
MEFNet~\cite{ma2019deep} used guided filter~\cite{he2012guided} to realize upsampling, which can well preserve the high-frequency and edge information. However, this method is too complicated and time-consuming even on small resolution images. Therefore, we propose a new MEF method based on LUT for the efficient and high quality fusion.  


Another problem faced by MEF task on mobile devices is that there lack of datasets captured by causal cameras, although there exist public datasets such as SICE~\cite{cai2018learning} and HDREYE~\cite{nemoto2015visual} taken by professional cameras, which means the images is high-quality. The trained model by these datasets may be hard to be applied to mobile devices due to the generalization problem. Therefore, we propose a comprehensive dataset to broaden the applications.

\subsection{LUT}\vspace{-0.5mm}
The LUT has been widely used in vision tasks, including image enhancement~\cite{zeng2020learning, wang2021real}, super-resolution~\cite{jo2021practical, li2022mulut} and so on. ~\cite{zeng2020learning, wang2021real} proposed image-adaptive 3D LUTs for efficient single image enhancement and they all need a network weight predictor to fuse different 3D LUTs, which will be restricted for some platforms that require deep learning framework support. ~\cite{jo2021practical} train a deep super-resolution (SR) network with a restricted receptive field and then cache the output values of the learned SR network to the LUTs. Compared with~\cite{zeng2020learning, wang2021real}, MEFLUT and SR-LUT~\cite{jo2021practical} are offline LUTs. In addition, MEFLUT is also essentially different from SR-LUT~\cite{jo2021practical} in the form of generating LUT and training strategy.


\section{Algorithm}\label{sec:algo}
Our method is divided into two stages, where Stage 1 is training a network $\mathcal{H}(\cdot)$ to estimate weight maps $\{W_k\}$ for a given set of $K$ YUV input images $\{I^L_k\}, k=1,...,K$, and Stage 2 is generating $K$ 1D lookup tables (LUTs) of size 256, composing a $K\times256$ matrix $\textbf{L}\in \mathbb{R}^{K\times 256}$, to achieve fast calculation of $\{W_k\}$ by simply querying $\textbf{L}$ during deployment. With $\{W_k\}$ obtained, the output image $I^H$ is obtained by alpha blending $\{I^L_k\}$ and $\{W_k\}$.
\subsection{Network Structure}\label{sec:ns}
Our network $\mathcal{H}(\cdot)$ is built upon DNN with two newly proposed modules, which aim at inter-frame feature fusion and intra-frame weight prediction by attention mechanism in various dimensions, respectively. It takes $K$ YUV images $\{I^L_k\}$ of size $H\times W$ as input, predicts $K$ weight maps $\{W_k\}$ of the same size as intermediate results, and finally produces the output image $I^H$ by alpha blending $\{I^L_k\}$ using $\{W_k\}$.
Note that our network only involves computation of Y channel as it is sufficient to predict the weight maps as a common experience, while UV channels are separately merged using a simple weighted summation as in~\cite{ram2017deepfuse}.

\begin{figure*}[t!]
\vspace{-6mm}
 \centering
  \begin{minipage}{8cm}
   \centering
   \includegraphics[width=0.90\linewidth]{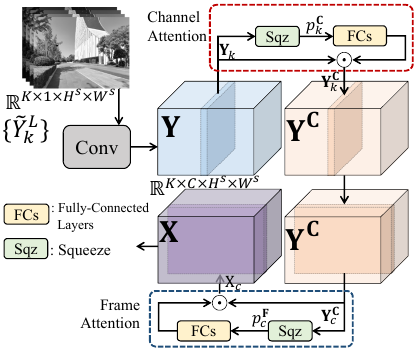}
  \end{minipage}
    \vspace{1mm}
 \begin{minipage}{8cm}
  \centering
  \includegraphics[width=0.90\linewidth]{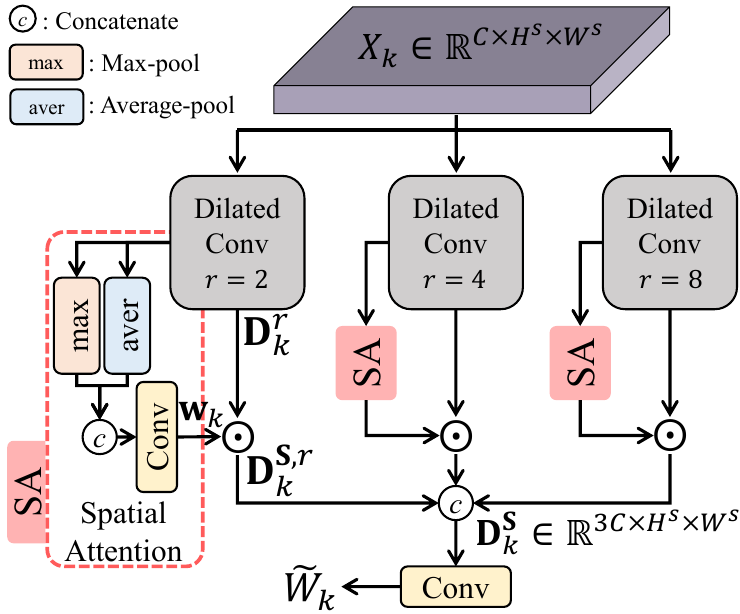}
 \end{minipage}
\caption{\textbf{Left}: Convolution with frame and channel attention (CFCA). \textbf{Right}: Dilated inception with spatial attention (DISA).}
\label{fig:att_module}
\vspace{-3mm}
\end{figure*}
\vspace{-2.0mm}
\subsubsection{Convolution with frame and channel attention}\label{sec:fffca}
The Y channel of the input frames $\{I^L_k\}$, i.e. $\{Y^L_k\}\in\mathbb{R}^{1\times H\times W}$ are first bilinearly downsampled at rate $s$ to obtain
$\{\widetilde{Y}^L_k\}$ with $\widetilde{Y}^L_k\in\mathbb{R}^{1\times H^s \times W^s}$. Then they are fed into a layer composed of convolution with frame and channel attention~\cite{hu2018squeeze} (CFCA) to obtain a
feature map $\textbf{X}$. Specifically, we first separately apply conv2d to each $\widetilde{Y}^L_k$ and concatenate the results together to get a 4D tensor $\textbf{Y} \in \mathbb{R}^{K\times C\times H^s\times W^s}$, and then we successively apply attention mechanism in the frame and channel dimension as,\\
\noindent 1) Channel attention:\vspace{-2.1mm}
\begin{small}
\begin{equation}\label{eq:maca}
\begin{aligned}
p^\textbf{C}_{k,c}&=\frac{1}{H^s\times W^s}\sum^H_{i=1}\sum^W_{j=1}\textbf{Y}_{k,c}(i, j),\\
\textbf{Y}^\textbf{C}_k &= \textbf{Y}_{k} \odot \sigma\big(\textbf{W}^\textbf{C}_2\cdot\delta(\textbf{W}^\textbf{C}_1\cdot p^\textbf{C}_k)\big),\\
\textbf{Y}^\textbf{C} &= \{\textbf{Y}^\textbf{C}_k\},\quad k=1,...,K. 
\end{aligned}
\end{equation}
\end{small} \vspace{-1mm}
\noindent 2) Frame attention:\vspace{-1.5mm}
\begin{small}
\begin{equation}\label{eq:mafa}
\begin{aligned}
p^\textbf{F}_{k,c} &=\frac{1}{H^s\times W^s}\sum^H_{i=1}\sum^W_{j=1}\textbf{Y}^\textbf{C}_{k,c}(i, j), \\
\textbf{X}_c &= \textbf{Y}^\textbf{C}_{c} \odot \sigma\big(\textbf{W}^\textbf{F}_2\cdot\delta(\textbf{W}^\textbf{F}_1\cdot p^\textbf{F}_c)\big), \\
\textbf{X} &= \{\textbf{X}_c\},\quad c = 1,...,C,
\end{aligned}
\end{equation}
\end{small}
where $p^\textbf{C}_{k} \in \mathbb{R}^{C\times 1\times 1}$, $p^\textbf{F}_{c} \in \mathbb{R}^{K\times 1\times 1}$ are vectors composed of scalars $p^\textbf{C}_{k,c}$ and $p^\textbf{F}_{k,c}$, respectively. $\textbf{W}^\textbf{C}_1,\textbf{W}^\textbf{C}_2,\textbf{W}^\textbf{F}_1,\textbf{W}^\textbf{F}_2$ are linear weights in two attention modules, and $\delta, \sigma$ are ReLU and sigmoid activation. $\odot$ is element-wise product with broadcasting automatically.

After the above steps, the two attention modules fuse the feature in frame and channel dimensions separately, so that the produced feature map $\textbf{X}$ accumulates rich context to predict the weight maps $\{\widetilde{W}_k\}$ in the following steps. As seen in Fig.~\ref{fig:cfca-disa-vis}, by comparing (e) and (c), or (d) and (b), we observe that with CFCA enabled, the weight map of the sky area (in green box) in EV-2 is highlighted, causing the final output image not over-saturated. This phenomenon proves the attention focuses more on EV-2 with CFCA involved. We illustrate CFCA structure in Fig.~\ref{fig:att_module}(\textbf{Left}).
\vspace{-2.0mm}
\subsubsection{Dilated inception with spatial attention}\label{sec:wpdisa}
\begin{figure}[t]
  \vspace{-6mm}
  \centering
  \includegraphics[width=1.0\linewidth]{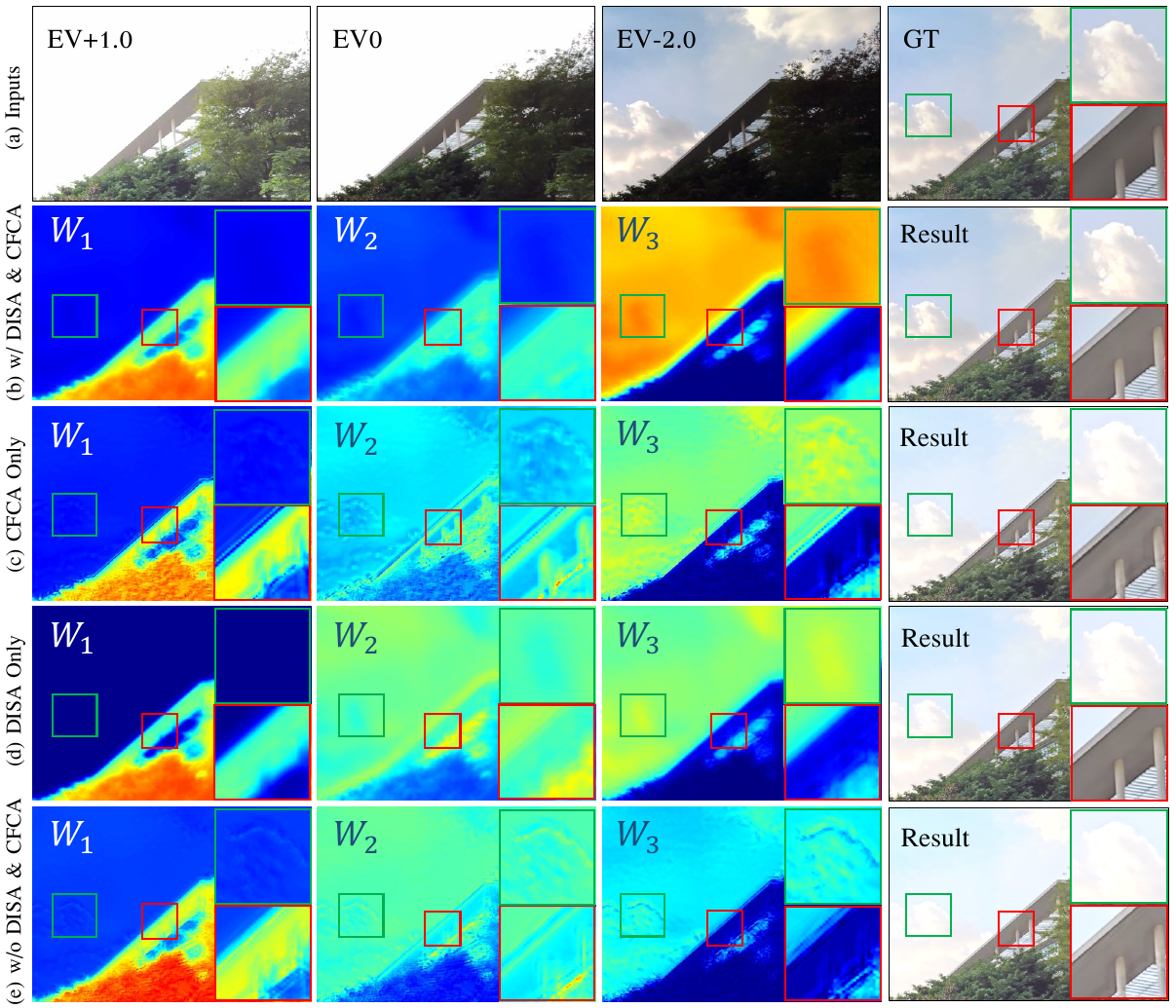}
  \caption{Visualization of weight maps produced by various cases. (a) Inputs and the GT target. From (b) to (e), weight maps of the 3 frames (column 1 to 3) and the final fusion image (column 4), for various cases. For weight maps, color map from deep blue to red represents value range $[0,1]$. }\vspace{-4mm}\label{fig:cfca-disa-vis}
\end{figure}

We further split $\textbf{X}$ in frame dimension into $X_1,X_2,...,X_K~(X_k\in\mathbb{R}^{C\times H^s\times W^s})$, and for each $X_k$, we feed it into a dilated inception layer with spatial attention (DISA). This layer consists of 3 branches of dilated convolution in rate $r = 2,4,8$ respectively, each of which produces a feature map with spatial attention~\cite{woo2018cbam} being applied to. Mathematically, the process is\vspace{-2mm}
\begin{small}
\begin{equation}\label{eq:disa}
\begin{aligned}
\textbf{D}^r_k &= \mathcal{D}(X_k, r) \in\mathbb{R}^{C\times H^s\times W^s}, \qquad r \in \{2,4,8\}, \\
\textbf{w}_k &= \sigma\Big(\Big[\frac{1}{C}\sum_{c=1}^C\textbf{D}^r_k(c),~\max_c \textbf{D}^r_k(c)\Big] \circledast \textbf{W}^\mathcal{D}\Big)\\
&~~~~~~~\qquad\qquad\textbf{D}^{\textbf{S},r}_k = \textbf{D}^r_k \odot \textbf{w}_k,
\end{aligned}
\end{equation}
\end{small}
where $\mathcal{D}(\cdot,r)$ is dilated conv2d with rate $r$, $[\cdot,\cdot]$ is concatenation and $\circledast$ is conv2d operator. With $\textbf{D}^{\textbf{S},r}_k$ in 3 branches obtained, we concatenate them together and apply another convolution so as to produce the final weight map $\widetilde{W}_k$. With DISA involved, the weights for pixels within a frame is spatially optimized so that details can be well preserved and artifacts are suppressed. For example in Fig.~\ref{fig:cfca-disa-vis}, we observe that the weights in red boxes become smoother from (e) to (d) or from (c) to (b), where (d)(b) additionally involve DISA based on (e)(c). 

Having completed the above steps, we obtain a multi-scale spatial attention from different dilated rates, so that the produced feature map obtain finer spatial structural information to predict the weight maps $\{\widetilde{W}_k\}$ in the following steps. The detailed structure of DISA is illustrated in Fig.~\ref{fig:att_module}(\textbf{Right}). $C$ is set to $24$ in all modules, and we make sure $\min\{H^s, W^s\} = 128$ regardless of the input image size by controlling $s$.

We conduct an ablation study to demonstrate the effectiveness of the two attention modules, where with either module enabled, all metrics become higher.
\vspace{-2mm}
\subsubsection{Unsupervised learning of the MEFLUT}
With $\{\widetilde{W}_k\}$ obtained, we further learn a high-quality detailed image in an unsupervised manner.
\vspace{1mm}\textit{\\Guided filtering for upsampling (GFU).}
We first apply guided filtering for upsampling (GFU)~\cite{ma2019deep} to resize $\{\widetilde{W}_k\}$ back to $\{W_k\}$ with its original resolution $H\times W$. This process also relies on $\{\widetilde{Y}^L_k\}$ and $\{Y^L_k\}$ which provide guiding information to restore visual pleasing weight maps. These weight maps further alpha blends the Y channels $\{Y^L_k\}$ to achieve $Y^H$, i.e.\vspace{-2mm}
\begin{small}
\begin{equation}
Y^H = \sum_{k=1}^K W_k \cdot Y^L_k.
\end{equation}
\end{small}
With GFU involved, higher quality output can be achieved than using bilinear upsampling.
\vspace{1mm}\textit{\\Loss function.}
As normally ground-truth detailed images are rarely available, we apply the MEF-SSIM loss in~\cite{ma2019deep} to train our network in an unsupervised way, i.e. \vspace{-1mm}
\begin{small}
\begin{equation}
\min_\theta \mathcal{L}_\text{MEF-SSIM}(\{Y^L_k\}, Y^H),
\end{equation}
\end{small}
where $\theta$ represents all the parameters in our network.
\vspace{1mm}\textit{\\U/V channels' merging.}
For U/V channels, we adopt the weighted sum method in~\cite{ram2017deepfuse} to merge various frames into one, i.e.\vspace{-1.0mm}
\begin{small}
\begin{align}
P^H = \frac{\sum_{k=1}^K ||P_k - \tau||^1_1\cdot P_k}{\sum_{k=1}^K ||P_k - \tau||^1_1},~~P\in\{U,V\},
\end{align}
\end{small}
where $\tau = 128$. With all channels obtained, the output $I^H$ is composed.

\subsection{LUT Generation}\label{sec:lut}
Unlike other image algorithms which are sufficiently deployed in PC or cloud platform, MEF commonly works in mobile platform and it requires extreme efficiency. To achieve it, we further involve a LUT matrix $\textbf{L} \in \mathbb{R}^{K\times 256}$ to speed up the inference procedure, with its $k$-th row corresponding to a 1D LUT for the $k$-th exposure image. To build up $\textbf{L}$, we feed the trained network $\mathcal{H}(\cdot)$ with 256 groups of constant grayscale images, with grayscale varying from 0 to 255. For example, for the $(v+1)$-th image group, we set $\{\widetilde{Y}^L_k\}(i,j) = v, \forall v \in \{0,1,...,255\}$ where $(i,j)$ is a pixel location. This constant grayscale image group $\{\widetilde{Y}^{L,v}_k\}$ produces weight maps $\{\widetilde{W}^v_k\}$ which are further frame-wisely averaged into $K$ scalars $\{\omega^v_k\}$, i.e. $\omega^v_k = \text{Mean}(\widetilde{W}^v_k)$. We set LUT matrix $\textbf{L}$ by \vspace{-1.8mm}
\begin{small}
\begin{align}\label{eq:lut-fill}
\textbf{L}(k,v+1) = \omega^v_k, \quad k=1,...,K; v=0,...,255,
\end{align} 
\end{small} 
so that all the entries are filled after 256 groups of images are fed. During deployment, given a set of $K$ input images $\{Y^L_k\}$, instead of feeding the network $\mathcal{H}$ again, we directly query $\textbf{L}$ for each pixel location $(i,j)$ where $v^*=Y^L_k(i,j)$, and set the weight map value by\vspace{-1.0mm}
\begin{small}
\begin{align}\label{eq:lut-query}
\widetilde{W}_k(i,j) = \textbf{L}(k,v^*+1).
\end{align} 
\end{small}
 For a 4K image, this strategy makes the time cost around 3ms which leaves enough room for mobile platform deployment, and ensures our method outperforms the others in efficiency. For example, inference with our network directly costs 16ms in NVIDIA 2080Ti GPU. Detailed comparison data can be found in Fig.~\ref{fig:time} and Tab.~\ref{table:diff_frame_time}. We also visualize a LUT in Fig.~\ref{fig:lut_vis} for clearer illustration.

\begin{figure}[t]
\vspace{-6mm}
   \centering
   \includegraphics[width=0.93\linewidth]{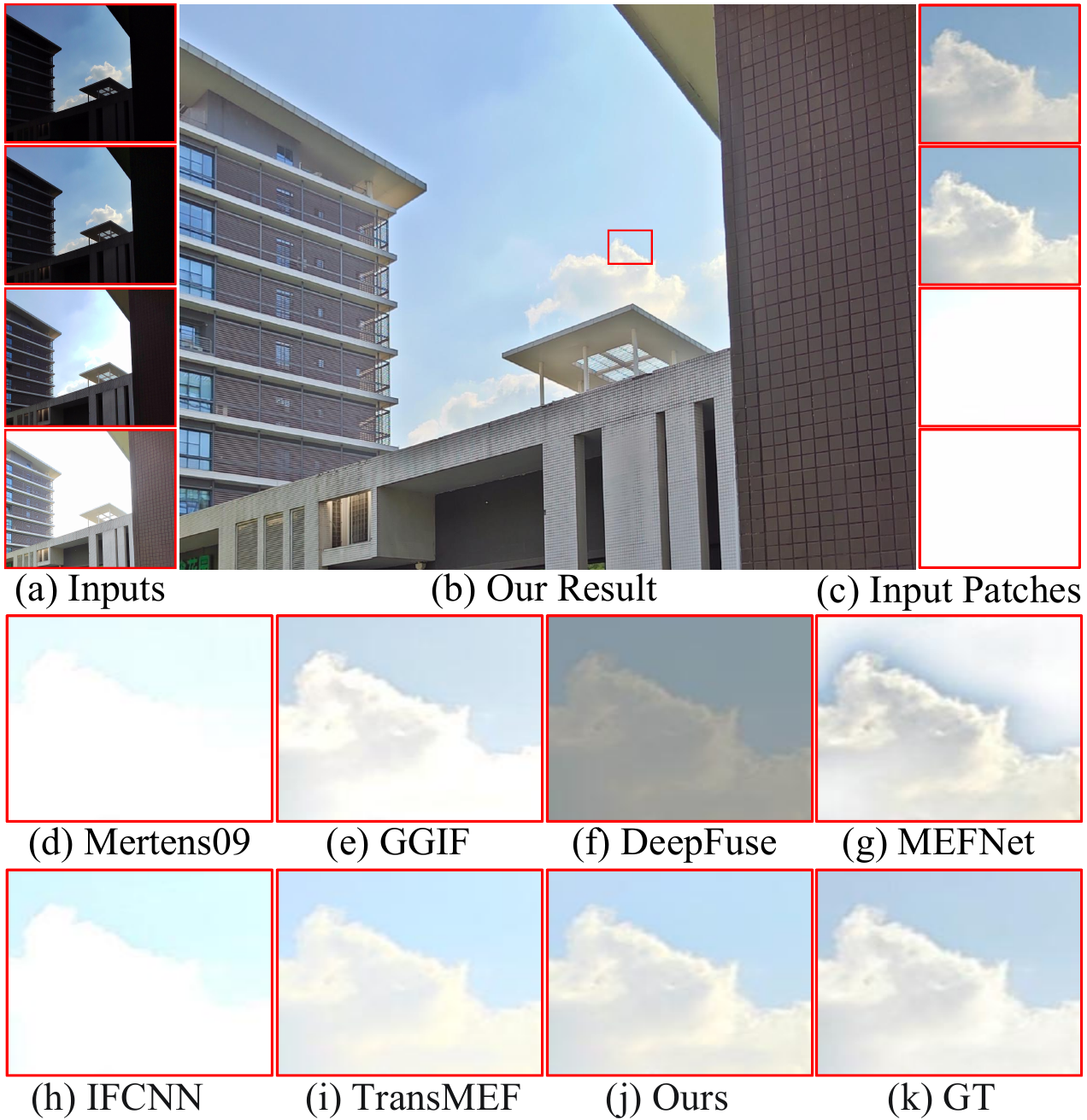}\vspace{-1mm}
   \caption{
  Qualitative comparison of (j) our method with (d) Mertens09, (e) GGIF, (f) DeepFuse, (g) MEFNet, (h) IFCNN and (i) TransMEF on our test set. (a) is inputs. 
   } \label{fig:our_vis}\vspace{-3mm}
 \end{figure}

\section{Data Preparation}\label{sec:dp}
\vspace{-1mm}
Considering there are few existing datasets of multi-exposure image sequences from mobile phones, and the provided image sequences as in~\cite{kang2003high, cai2018learning, nemoto2015visual} are very limited in quantity and diversity, we create a new dataset that contains higher number of multi-exposure image sequences and covers more diverse scenes captured by mobile phones.


\begin{figure*}[htbp]
\vspace{-6mm}
\centering
\includegraphics[width=0.95\linewidth]{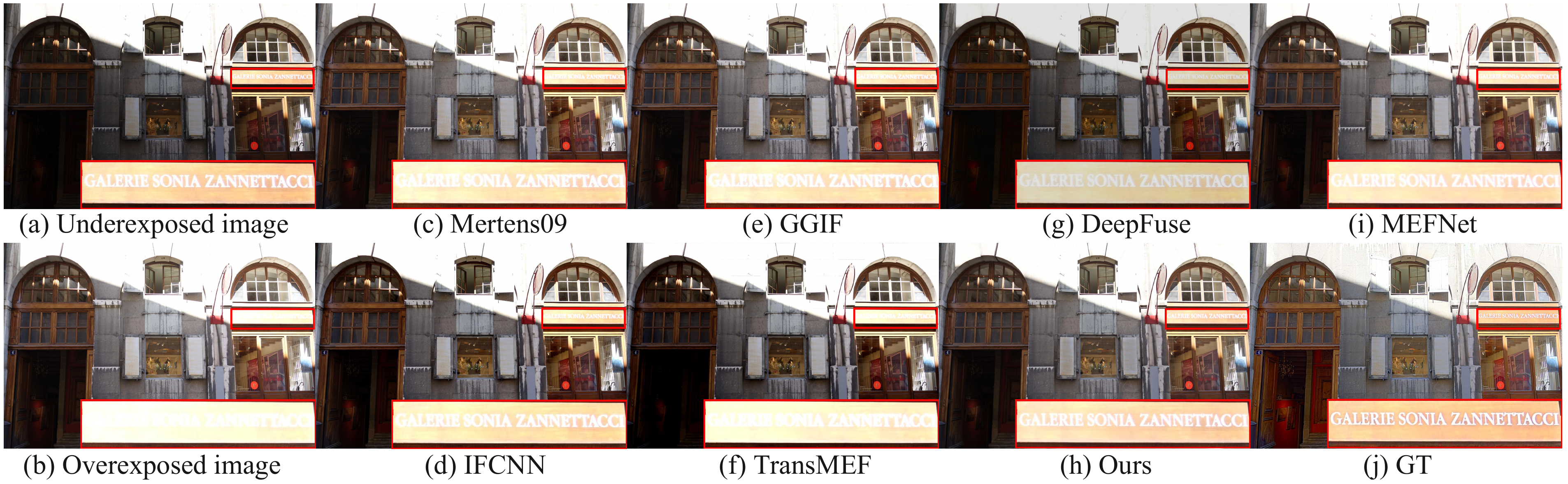}\vspace{-1mm}
\caption{
Qualitative comparison of (h) our method with (c) Mertens09, (e) GGIF, (g) DeepFuse, (i) MEFNet, (d) IFCNN and (f) TransMEF on SICE dataset. Input sequences with different exposures are shown in (a) and (b). 
} \label{fig:sice_vis}\vspace{-4mm}
\end{figure*}

\vspace{-3mm}
\paragraph{Data collection.}
We collect the data mainly in static scenes using 6 different commonly used brands of mobile phones. The scenes are ensured diverse and representative, which cover a broad of scenarios, subjects, and lighting conditions. More importantly, the collected images cover most of the exposure levels in our daily life. For each sequence, we use a tripod to ensure the frames are well-aligned. The exposure levels are manually set, and Exposure Values (EVs) of our sample sequences setting range from $-4.0$ to $+2.0$ with $0.5$ as a step. We select the exposure number 6 ($K=6$) for each scene based on the characteristics of different brands of mobile phones. After collecting the source sequences, screening is further conducted to select desirable sequences for GT generation.
As a result, a total of 960 static but diverse sequences are filtered out.

\vspace{-3mm}
\paragraph{GT generation.}
We further use a hybrid method to generate the GT. Specifically, 14 existing algorithms~\cite{mertens2009exposure,raman2009bilateral,shen2011generalized,zhang2011gradient,li2013image,shen2014exposure,ma2017robust,kou2017multi,li2020fast,sen2012robust,hu2013hdr,bruce2014expoblend,oh2014robust,ma2019deep} were first used to predict results for each sequence. Then we invited 20 volunteers to compare the results of 14 algorithms and vote for an image as the GT for each sequence. We also invited an image quality tuning engineer to further manually tune the tone-mapping operators for the fairly voted results to generate the high-quality GT. The average tuning time for each image is 10 minutes, and each volunteer spent over 60 minutes for the entire voting. Due to the huge workload, we tuned and obtained high-quality GT on the 155 samples of test set only to evaluate the results by unsupervised learning, which costs totally $2450/60=40.8$ man-hours.

\begin{table*}[t]
    \label{tab:psnr_noniid}
    \begin{center}
    \vspace{-6mm}
    
    \resizebox{0.95\linewidth}{!}{
    \setlength{\tabcolsep}{0.3 pt}
    \begin{tabular}{
    >{\centering\arraybackslash}p{3.2cm}
    >{\centering\arraybackslash}p{1.9cm}
    >{\centering\arraybackslash}p{2.2cm}
    >{\centering\arraybackslash}p{1.5cm}
    >{\centering\arraybackslash}p{1.5cm}
    >{\centering\arraybackslash}p{1.5cm}
    >{\centering\arraybackslash}p{2.0cm}
    >{\centering\arraybackslash}p{2.0cm}
    >{\centering\arraybackslash}p{1.5cm}
    >{\centering\arraybackslash}p{1.5cm}
    >{\centering\arraybackslash}p{1.5cm}
    }
         \toprule
         Exposure Number ($K$)&Metrics &  Mertens09 &  Li13   &GGIF & Li20 & DeepFuse & MEFNet & IFCNN & TransMEF & Ours \\
        \hline
        \multirow{4}*{$K=2$} & PSNR$\uparrow$
                    & 26.886 & 19.475 & 27.172  & 25.997  &23.087 & 28.763 &29.085 & \textcolor{blue}{29.285} & \textcolor{red}{29.443}\\
          &  SSIM$\uparrow$  & 0.8347 &0.8326   &0.9153  &0.8765  &0.7558     & 0.9468   & 0.9373& \textcolor{blue}{0.9568} & \textcolor{red}{0.9587} \\
          & Q$_\emph{C}$ $\uparrow$  &  0.5531 &0.5166 & 0.5330 & 0.5245 &0.4603     & 0.6596    & 0.6322 & \textcolor{blue}{0.6796} & \textcolor{red}{0.6860}\\
          &  MEF-SSIM$\uparrow$  & 0.9410 &0.8797 &0.9638 & 0.9273 &0.7407     & 0.9633    & \textcolor{blue}{0.9683} & 0.9663& \textcolor{red}{0.9773}\\
        \hline
        \multirow{4}*{$K=3$} & PSNR$\uparrow$
                    & 27.328 & 20.922  & 29.437 & 25.396 & -  &   29.793 & \textcolor{blue}{30.291} & - & \textcolor{red}{31.387}\\
          &  SSIM$\uparrow$  &0.8927 &0.8468   &0.9208  &0.8603  &  -        & \textcolor{blue}{0.9569}   &  0.9473 & - & \textcolor{red}{0.9590} \\
          &  Q$_\emph{C}$ $\uparrow$  & 0.4779 &0.4562 &0.4735  & 0.4742 &  -      &\textcolor{blue}{0.6054}   & 0.5798 & - & \textcolor{red}{0.6075}\\
          &  MEF-SSIM$\uparrow$  & 0.9452 &0.8955 &0.9592  & 0.9387 &  -      &0.9593   & \textcolor{blue}{0.9612} & - & \textcolor{red}{0.9688}\\
        \hline
        \multirow{4}*{$K=4$} & PSNR$\uparrow$
                    & 26.154 & 20.471   &27.782  &24.602  & - & 29.365    & \textcolor{blue}{29.872}& - & \textcolor{red}{31.025}\\
        &  SSIM$\uparrow$    & 0.8834 &0.8276   &0.9087  &0.8448  & -      & \textcolor{blue}{0.9513}   & 0.9501 & - & \textcolor{red}{0.9548}\\
        &  Q$_\emph{C}$ $\uparrow$    & 0.4379 & 0.4107   &0.4359  & 0.4371 & -     & \textcolor{blue}{0.5493}     & 0.5421 & - &  \textcolor{red}{0.5571}\\

        &  MEF-SSIM$\uparrow$    & 0.9360 &0.8897   &0.9528  & 0.9096 & -     & 0.9512     & \textcolor{blue}{0.9543} & - &  \textcolor{red}{0.9603}\\
        \bottomrule
        \multicolumn{11}{c}{(a) Results on our dataset.}
    \end{tabular}}\\
    \resizebox{0.95\linewidth}{!}{
      \setlength{\tabcolsep}{7 pt}
    \begin{tabular}{
    >{\centering\arraybackslash}p{1.9cm}
    >{\centering\arraybackslash}p{1.5cm}
    >{\centering\arraybackslash}p{1.5cm}
    >{\centering\arraybackslash}p{1.5cm}
    >{\centering\arraybackslash}p{1.5cm}
    >{\centering\arraybackslash}p{1.5cm}
    >{\centering\arraybackslash}p{1.5cm}
    >{\centering\arraybackslash}p{1.5cm}
    >{\centering\arraybackslash}p{1.5cm}
    >{\centering\arraybackslash}p{1.5cm}}
        \toprule
        Metrics & Mertens09 &  Li13   &GGIF & Li20 &DeepFuse &MEFNet & IFCNN & TransMEF  & Ours \\
        \hline
         PSNR$\uparrow$
                    & 21.531 & 19.380   & 21.316 &21.072 &  20.915 & 21.583  & \textcolor{blue}{21.626} & 21.601 & \textcolor{red}{21.894}\\
          SSIM$\uparrow$  & 0.7833 &0.7732   &0.7929  &0.7743  &0.6872     & 0.7763  &0.7839  & \textcolor{blue}{0.7965} & \textcolor{red}{0.8067} \\
          Q$_\emph{C}$ $\uparrow$  & 0.7759 &0.7056  &0.7709 & 0.7747 &0.6406     & 0.7760  & 0.7789&  \textcolor{blue}{0.7837}& \textcolor{red}{0.7934}\\
          MEF-SSIM$\uparrow$  & 0.9589 &0.9451  &0.9692 & 0.9498 &0.8590     & 0.9743  & 0.9756 & \textcolor{blue}{0.9782} & \textcolor{red}{0.9845}\\

        \bottomrule
        \multicolumn{10}{c}{(b) Results on SICE~\cite{cai2018learning} dataset.}
    \end{tabular}}

    \end{center}
    \vspace{-3mm}
    \caption{The PSNR, SSIM, Q$_\emph{C}$, MEF-SSIM of all methods on our test set (a) and SICE dataset (b). The best and 2nd best results are highlighted in \textcolor{red}{red} and in \textcolor{blue}{blue}, respectively. `-' means not applicable, as DeepFuse and TransMEF can only take two images as input.}
     \label{table:ourd}
    \vspace{-3mm}
\end{table*}

\section{Experimental Results}

\subsection{Implementation Details}\vspace{-1mm}
\paragraph{Dataset.}
We conduct experiments on our dataset, and another public static dataset SICE~\cite{cai2018learning} is involved to keep the comparison fair enough. In our dataset, a total of 960 samples, of which 805 samples are used for training and 155 samples are used for evaluation. For SICE, we just utilized its first part of 360 index-available sequences with more classic pictures the authors obtained from a widely used dataset HDREYE~\cite{nemoto2015visual}. Following the setting of SICE, 302 samples for training, remaining 58 samples for evaluation.
\vspace{-3mm}
\paragraph{Evaluation.}
We use PSNR, SSIM, and another unsupervised metric Q$_\emph{C}$~\cite{cvejic2005similarity} as the main evaluation metric for all the methods we compare, besides the training loss MEF-SSIM~\cite{ma2015perceptual}.
Also, considering most of the involved methods are unsupervised ones, causing a large difference in brightness that may exist in their results, we calculate PSNR after an average brightness is subtracted from the given image for fair comparison.

\vspace{-3mm}
\paragraph{Training.}
We fix the downsampling rate $s$ to $4$ so that the input training image are fixed size of $512^2$. We use the ADAM optimizer~\cite{Adam} with momentum terms (0.9, 0.999), and the learning rate is set to $1e^{-4}$. The total training epoch is $100$. Finally, we evaluate our method at full resolution during testing.

\subsection{Comparison with Existing Methods}\vspace{-1mm}

We compare our method with 8 previous MEF methods on our test set and SICE test set, including Mertens09~\cite{mertens2009exposure}, Li13~\cite{li2013image}, GGIF~\cite{kou2017multi}, Li20~\cite{li2020fast}, DeepFuse~\cite{ram2017deepfuse}, MEFNet~\cite{ma2019deep}, IFCNN~\cite{zhang2020ifcnn}, and TransMEF~\cite{qu2022transmef}. The results of each method are generated by the implementations from the original authors with default settings.

\vspace{-3mm}
\paragraph{Qualitative comparison.}
Fig.~\ref{fig:our_vis} shows the visual comparisons of fused images generated through other methods on our test set. As seen, the results of the other methods commonly fail to recover the details of saturated regions in the sky. In contrast, our method hallucinates more accurate contents in over saturated areas, causing the sharp edge of the cloud well preserved. Fig.~\ref{fig:sice_vis} shows the visual comparisons on SICE dataset. As seen, the red box highlights that other methods produce a color cast in the billboard area without recovering the text details, while in contrast, our method produces a natural appearance with faithful detail and color restoration.




\vspace{-3mm}
\paragraph{Quantitative comparison.}
As shown in Tab.~\ref{table:ourd}(a), we conduct experiments on three sub-datasets with exposure numbers set to 2, 3, and 4, we select the exposures we need from each sample at multiple fixed EVs intervals. As seen from the table, our method consistently achieves the best results over the others in various exposure numbers. We also evaluate all methods on SICE. Here we conduct experiments with the exposure number being set to 2 (EV-1, EV+1), this setting follows the index-available sequences provided by SICE. As seen from the Tab.~\ref{table:ourd}(b), our method achieves similar results, verifying the generalization of our method.
\begin{figure}[t]
   \centering\vspace{-3mm}
   \includegraphics[width=0.95\linewidth]{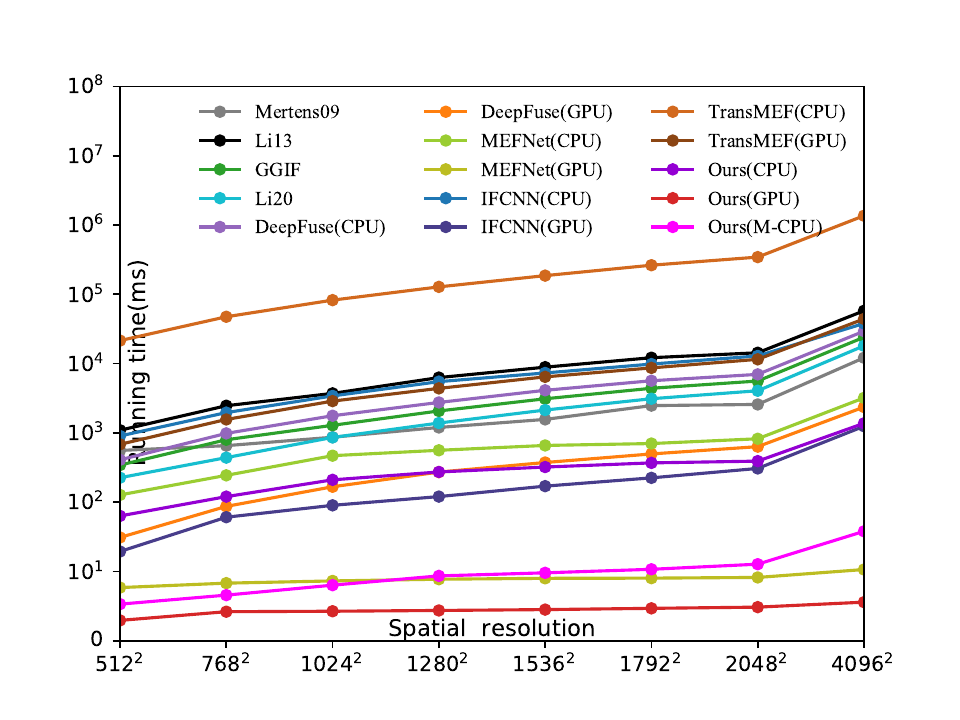}
   \caption{Running time in milliseconds (ms) on different spatial resolutions ($K=2$).} \label{fig:time}\vspace{-4mm}
 \end{figure}

\begin{table}[t]
	\centering
	\vspace{-2mm}
	\setlength{\tabcolsep}{6 pt}
\resizebox{0.95\linewidth}{!}{
\begin{tabular}{
>{\centering\arraybackslash}p{1.0cm}
>{\centering\arraybackslash}p{1.3cm}
>{\centering\arraybackslash}p{1.7cm}
>{\centering\arraybackslash}p{1.7cm}
>{\centering\arraybackslash}p{1.7cm}
}
\toprule
                     &\multirow{2}*{Methods}& \multicolumn{3}{c}{Exposure Number ($K$)} \\
                     \cline{3-5}
                     & & 2 & 3 & 4                 \\ \hline
\multirow{9}{*}{CPU} & Mertens09 & 2575                       & 3570                    & 4624                        \\
                     & Li13       & 14336                       & 17708                      & 20479                      \\
                     & GGIF      & 5605                        & 8268                       & 10861                       \\
                     & Li20      & 4057                        & 5973                        & 7835                        \\
                     & DeepFuse  & 6991                        & -                                 & -                              \\
                     & MEFNet    & \textcolor{blue}{822}(14833)                        & \textcolor{blue}{861}(15097)                     & \textcolor{blue}{890}(15641)                        \\
                      & IFCNN    & 12901                        & 18123                     & 23954                        \\%
                       & TransMEF    & 344168                        & -                     & -                        \\%
                     & Ours      & \textcolor{red}{391}(54)       &\textcolor{red}{413}(56)                          & \textcolor{red}{435}(60)                        \\ \hline
\multirow{5}{*}{GPU}  & DeepFuse  & 634                          & -                              & -                              \\
          & MEFNet    & \textcolor{blue}{8.21}(5.48) & \textcolor{blue}{8.65}(5.92) & \textcolor{blue}{8.72}(6.31) \\ 
           & IFCNN    & 307 & 360 & 417 \\ 
           & TransMEF    & 11524 & - & - \\ 
         & Ours      & \textcolor{red}{3.06}(0.53) & \textcolor{red}{3.54}(0.61) & \textcolor{red}{3.89}(0.65) \\
         \hline
\multirow{1}{*}{M-CPU} 
         & Ours      & 11.78 & 12.75 & 13.32 \\
\bottomrule
\end{tabular}}
\vspace{1mm}
\caption{Running time of MEF methods on different $K$ settings in milliseconds (ms). `-' means not supported by DeepFuse and TransMEF for such a test. For values in `()', we only compare our method with MEFNet, they are calculated by disabling GFU so that the speed of LUT only can be directly measured. M-CPU denotes the mobile CPU. The best and 2nd best are marked in \textcolor{red}{red} and \textcolor{blue}{blue}, respectively.}\vspace{-4mm}
\label{table:diff_frame_time}
\end{table}

\vspace{-3mm}
\paragraph{Running time.}

We conduct a speed comparison of our method with the 8 MEF methods on our test set, in various platforms including CPU, GPU and mobile CPU, as shown in Tab.~\ref{table:diff_frame_time}. There, CPU is an Intel i7-10510U 1.8GHz, GPU is an Nvidia GeForce RTX 2080Ti and the mobile CPU is Qualcomm SM8250. Mertens09, Li13, GGIF and Li20 utilize CPU only, while DeepFuse, MEFNet, IFCNN and TransMEF exploit CPU and GPU, and ours exploit all three. We only compared the running time of DeepFuse and TransMEF on 2 exposures due to its strict input constraints. 

When the resolution is fixed to $2048^2$ (2K), our method run in 3.06ms in GPU, which is faster than MEFNet's speed (8.21ms) and DeepFuse's speed (634ms), and the speed of CPU is also significantly faster than that of DeepFuse's, MEFNet's and other methods such as Mertens09, Li13, GGIF, Li20, IFCNN and TransMEF, these methods show a lot of performance consumption.
In mobile CPU, our method is able to achieve real-time performance. Compared with PC CPU, our method optimizes the performance of query and other operations for mobile CPU, and our method has no strict requirements on the running platform compared with deep learning methods such as Deepfuse, MEFNet, IFCNN and TransMEF. 

We also conduct experiments on images with various resolutions and illustrate the performance in Fig.~\ref{fig:time}. As seen, our method takes less than 4ms to process image sequences with resolutions ranging from $512^2$ to $4096^2$ on the GPU. 
The network of MEFNet downsamples the full resolution image to a fixed $128^2$, so their running time is relatively not much. Our method maintains the quality while achieving less running time, which is of $3.0\times$ and $175.0\times$ speed of MEFNet and DeepFuse, and $345.4\times$ and $12226.0\times$ speed of IFCNN and TransMEF in $4096^2$ resolution (4K). In CPU, our method is of $21.3\times$ and $42.0\times$ speed of DeepFuse and Li13, and $27.6\times$ and $989.6\times$ speed of IFCNN and TransMEF. In mobile CPU, our method running less than 40ms for a 4K image sequence. We further conduct an experiment by working on images in their original resolution by disabling the GFU module to verify the efficiency of our 1D LUTs only. The performance advantage of our luts is more obvious when GFU disabled.

Our method has been successfully shipped into millions of mobile phones in the market. This fact further demonstrates the practicality and robustness of our method.

\begin{table}[t]
  \vspace{-6mm}
	\centering
	\setlength{\tabcolsep}{6 pt}
	\resizebox*{0.95\linewidth}{!}{
\begin{tabular}{cc|cccc}
\toprule
CFCA         & DISA         & PSNR$\uparrow$            & SSIM$\uparrow$  & Q$_C$$\uparrow$  & MEF-SSIM$\uparrow$\\ 
          \hline
             &              & 29.648          & 0.9373   &$0.5617$    & 0.9442\\
$\checkmark$ &              & 30.570          & 0.9431   &$0.5943$  & 0.9522\\
             & $\checkmark$ & 30.782          & 0.9482   & $0.5960$  & 0.9598\\
$\checkmark$ & $\checkmark$ & \textbf{31.387} & \textbf{0.9590} & $ \textbf{0.6075}$ & \textbf{0.9688} \\
\bottomrule
\end{tabular}}
\vspace{1mm}
\caption{Ablation study of the proposed CFCA and DISA modules ($K=3$). }
\label{table:mod}
\vspace{0.6mm}
\resizebox{0.95\linewidth}{!}{
  \setlength{\tabcolsep}{10 pt}
  \begin{tabular}{c|cccc}
    \toprule 
    Methods  & PSNR$\uparrow$ & SSIM$\uparrow$ & Q$_C$$\uparrow$ & MEF-SSIM$\uparrow$ \\
    \hline
     MEFNet & 25.315 & 0.8924 & 0.5575 & 0.9064 \\
     Ours & \textbf{31.387} & \textbf{0.9590} & \textbf{0.6075} & \textbf{0.9688} \\
     
    \bottomrule
  \end{tabular}}
  \vspace{1mm}
  \caption{The Effectiveness of our network to LUTs ($K=3$).}
  \label{table:validity}
  \vspace{-3mm}
\end{table}

\subsection{Ablation Studies}



\begin{figure}[t]
   \centering\vspace{-7mm}
   \includegraphics[width=0.9\linewidth]{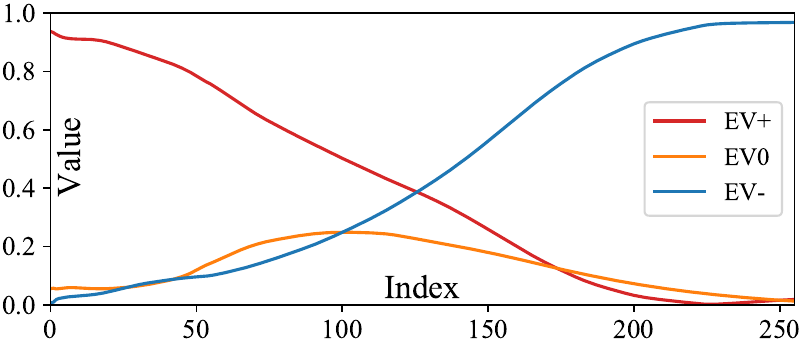}\vspace{-1mm}
   \caption{The visualization of the 1D LUTs ($K=3$).} \label{fig:lut_vis}\vspace{-1mm}
 \end{figure}

\paragraph{Visualization of LUTs.}
Fig.~\ref{fig:lut_vis} shows the weight value of the multi-exposure image in the range of 0 to 255 in 3 EVs. As seen, the curve of EV$+$ basically follows a monotonously decreasing pattern, while that of EV$-$ follows a monotonously increasing one. Our 1D LUTs reflect the fusion trend of different exposure images.
\vspace{-2mm}
\paragraph{The effectiveness of CFCA and DISA.}\label{par:effectiveness-cfca-disa}
We verify the module of CFCA and DISA respectively.
Tab.~\ref{table:mod} and Fig.~\ref{fig:cfca-disa-vis} shows that when the module of CFCA and DISA are also adopted, all metrics and visual effects can achieve the best performance. The main reason can be concluded as that the dual attention mechanism of the CFCA strengthens the exchange of information between different exposures, our 1D LUTs can better take into account the difference in brightness between different exposures, and the DISA acts on each branch and focuses on spatial features under different receptive fields, the consistency of spatial information can be preserved as much as possible.

\vspace{-2mm}
\paragraph{The effectiveness of our network to LUTs.}
In order to verify that the network we proposed is more effective for 1D LUTs' generation, we apply the same idea to MEFNet and used the generated 1D LUTs to perform the fusion task instead. As seen in Tab.~\ref{table:validity}, although MEFNet produces similar 1D LUTs to reconstruct the weight maps, its quality has a significant drop. It demonstrates that our network has better effectiveness for 1D LUTs' generation.

\begin{figure}[t]
\centering\vspace{-2mm}
\includegraphics[width=0.9\linewidth]{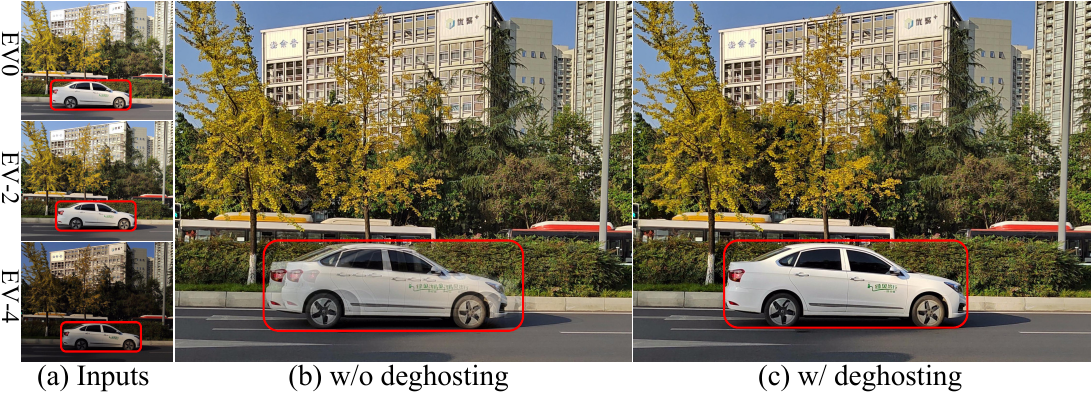}
\vspace{-1.5mm}
\caption{MEFLUT collaborates with a deghosting method. }
\label{fig:ghost}
\vspace{-3mm}
\end{figure}
\vspace{-2mm}
\paragraph{Moving scene.}
In our paper, we focus on MEF, similar to previous MEF works, such as DeepFuse, MEFNet, IFCNN and TransMEF, we all focus on static scenes. However, we also show how MEFLUT collaborates with other deghosting methods in practical applications. We show an example captured by a hand-held cellphone with dynamic objects in Fig.~\ref{fig:ghost}. We first align the input frames, then remove ghosts in the final image. As for solving the ghosts, we generate weight maps based on our 1D LUTs. The moving areas are detected and marked according to~\cite{zhang2015ghost}. The marked pixels are then assigned zero weights to prevent ghost artifacts. Our method can collaborate well with the off-the-shelf deghosting methods for practical results. 

\vspace{-2mm}
\begin{table}[t]
  \centering
  \resizebox{0.9\linewidth}{!}{
  \setlength{\tabcolsep}{2 pt}
  \begin{tabular}{c|ccccc}
    \toprule 
    Methods  & PSNR$\uparrow$ & SSIM$\uparrow$ & Q$_C$$\uparrow$ & MEF-SSIM$\uparrow$ & Running time (ms) \\
    \hline
     Network & \textbf{32.125} & \textbf{0.9674} & \textbf{0.6168} & \textbf{0.9764} & 12.37 \\
     LUTs & 31.387 & 0.9590 & 0.6075 & 0.9688 & \textbf{3.54} \\
    \bottomrule
  \end{tabular}}
  \vspace{1mm}
  \caption{Comparison experiment of our LUTs with our network. The running platform is PC GPU ($K=3$).}
  \label{table:lut_gap}
  \vspace{-5mm}
\end{table}
\paragraph{Trained network vs. LUTs.} 
Although the LUTs in our method do not participate in the training. As shown in Tab.~\ref{table:lut_gap}, we find that the quality of the results of the LUTs is comparable to the trained network, but the speed is several times faster. Our method has little effect in actual deployment, but using LUTs will get a large speed advantage with less quality loss. 

\subsection{Limitation and Discussion}
\paragraph{Limitation.} Although our method can achieve stable and efficient MEF across a variety of scenes, it has two limitations from 1D LUTs. First, our 1D LUTs currently only operates on the Y channel, which may not be good enough for color balance. One possible solution is to use the 1D LUTs on the UV channels as well but compatibility needs to be further investigated. Second, we query the 1D table pixel-wisely, causing the reconstructed weight map may be lack of smoothness and dot artifacts may appear as neighborhood information is not considered. Despite of GFU's smoothness capability on the weight map, its parameters cannot be adaptively set causing its uneven effects to various scenes, i.e. some are too dotted and some are over-smoothed resulting in Buddha light artifact. Two possible solutions include to learn a small model that provides more semantic information guidance for the 1D LUTs, and to learn a GFU with adaptive parameters.  We leave them as our future work. 
\paragraph{Discussion.}  The stronger the network's fitting capability, the greater the expressiveness of the generated LUTs. Advanced modules like Transformers and larger models can enhance the network's performance, resulting in superior 1D LUTs. Furthermore, our method accelerates MEF by offline generating 1D LUTs. It can be extended to tasks such as multi-focus image fusion and image enhancement. The same offline generation approach for 1D LUTs can also be applied to generate 2D/3D LUTs, enabling adaptation to a wider range of tasks. For instance, in a two-frame MEF task, we trained the network using  $256 \times 256$ groups of constant grayscale images, with grayscale values ranging from 0 to 255. We have successfully built  on offline-generated 2D LUTs exhibit promising performance. However, it is important to note that as LUT dimensionality increases, storage space requirements grow exponentially. We plan to further explore these possibilities in future research.



\section{Conclusion}\label{sec:conclu}
We have proposed a new method to efficiently fuse a multi-exposure image sequence to produce a visually pleasing result. We learn one 1D LUT for each exposure, then all the pixels from different exposures can query 1D LUT of that exposure independently for high-quality and efficient fusion. Specifically, to learn these 1D LUTs, we involve frame, channel and spatial attention mechanism into the MEF task to achieve superior performance.
We also release a new dataset composed of 960 multi-exposure image sequences collected from mobile phones of various brands and in diverse scenes.
We have conducted comprehensive experiments to demonstrate the effectiveness of MEFLUT.

\noindent\textbf{Acknowledgements} This work is supported by Sichuan Science and Technology Program under grant No.2023NSFSC0462.

{\small
\bibliographystyle{unsrt}
\bibliography{egbib}
}

\end{document}